\documentclass[a4paper,11pt]{article}

\usepackage{graphicx}  
\usepackage{url}       
\usepackage{times}
\usepackage{natbib}
\usepackage[margin=25mm]{geometry}
\usepackage{times}
\usepackage{latexsym}
\usepackage{url,setspace,comment}
\usepackage{verbatimbox,multirow,mathtools}

\title{Semantic Matching of Documents from\\Heterogeneous Collections:\\A Simple and Transparent Method for Practical Applications}  
\date{}

\author{Mark-Christoph M\"uller\\
       Heidelberg Institute for Theoretical Studies gGmbH\\
        Heidelberg, Germany\\
       \texttt{mark-christoph.mueller@h-its.org}
}

\begin{document}
\maketitle
\thispagestyle{empty}
\pagestyle{empty}

\begin{abstract}
We present a very simple, unsupervised method for the pairwise matching of documents from heterogeneous collections.
We demonstrate our method with the Concept-Project matching task, which is a binary classification task involving pairs of documents from heterogeneous collections. 
Although our method only employs standard resources without any domain- or task-specific modifications, it clearly outperforms the more complex system of the original authors. 
In addition, our method is \emph{transparent}, because it provides explicit information about how a similarity score was computed, and \emph{efficient}, because it is based on the aggregation of (pre-computable) word-level similarities. 
\end{abstract}

\section{Introduction}
\label{sec:introduction}
We present a simple and efficient unsupervised method for pairwise matching of documents from \emph{heterogeneous} collections. 
Following \cite{gong2018}, we consider two document collections heterogeneous if their documents differ systematically with respect to vocabulary and / or level of abstraction.
With these \emph{defining} differences, there often also comes a difference in length, which, however, by itself does not make document collections heterogeneous. 
Examples include collections in which \emph{expert} answers are mapped to \emph{non-expert} questions (e.g.\ \emph{InsuranceQA} by \cite{feng2015}),
but also so-called \emph{community} QA collections (\cite{blooma2011}), where the lexical mismatch between Q and A documents is often less pronounced than the length difference. 

\noindent
Like many other approaches, the proposed method is based on word embeddings as universal meaning representations, and on vector cosine as the similarity metric. 
However, instead of computing pairs of document representations and measuring their similarity, our method assesses the document-pair similarity on the basis of selected pairwise word similarities. 
This has the following advantages, which make our method a viable candidate for practical, real-world applications:
\textbf{efficiency}, because pairwise word similarities can be efficiently (pre-)computed and cached, and
\textbf{transparency}, because the selected words from each document are available as evidence for what the similarity computation was based on. 

\noindent
We demonstrate our method with the \emph{Concept-Project matching} task (\cite{gong2018}), which is described in the next section.

\section{Task, Data Set, and Original Approach}
\label{sec:task}
The \emph{Concept-Project matching} task is a binary classification task where each instance is a pair of heterogeneous documents:
one \textbf{concept}, which is a short science curriculum item from NGSS\footnote{\url{https://www.nextgenscience.org}}, and one \textbf{project}, which is a much longer science project description for school children from ScienceBuddies\footnote{\url{https://www.sciencebuddies.org}}. 
The publicly available data set\footnote{\url{https://github.com/HongyuGong/Document-Similarity-via-Hidden-Topics}} contains $510$ labelled pairs\footnote{Of the original $537$ labelled pairs, $27$ were duplicates, which we removed.} involving $C=75$ unique concepts and $P=230$ unique projects.
A pair is annotated as $1$ if the project matches the concept ($57\%$), and as $0$ otherwise ($43\%$).
The annotation was done by undergrad engineering students.
\cite{gong2018} do not provide any specification, or annotation guidelines, of the semantics of the 'matches' relation to be annotated. 
Instead, they create gold standard annotations based on a majority vote of three manual annotations. 
Figure \ref{fig:example} provides an example of a matching C-P pair.
The concept labels can be very specific, potentially introducing vocabulary that is not present in the actual concept descriptions.
The extent to which this information is used by \cite{gong2018} is not entirely clear, so we experiment with several setups (cf.\ Section \ref{sec:experiments}).

\begin{figure}
\begin{tiny}
\begin{spacing}{0.8}
\begin{quote}
\textbf{CONCEPT LABEL: ecosystems: - ls2.a: interdependent relationships in ecosystems}\\
\noindent
\textbf{CONCEPT DESCRIPTION:} Ecosystems have carrying capacities , which are limits to the numbers of organisms and populations they can support . These limits result from such factors as the availability of living and nonliving resources and from such challenges such as predation , competition , and disease . Organisms would have the capacity to produce populations of great size were it not for the fact that environments and resources are finite . This fundamental tension affects the abundance ( number of individuals ) of species in any given ecosystem .

\textbf{PROJECT LABEL: Primary Productivity and Plankton}\\
\textbf{PROJECT DESCRIPTION:} Have you seen plankton? I am not talking about the evil villain trying to steal the Krabby Patty recipe from Mr. Krab. I am talking about plankton that live in the ocean. In this experiment you can learn how to collect your own plankton samples and see the wonderful diversity in shape and form of planktonic organisms. The oceans contain both the earth's largest and smallest organisms. Interestingly they share a delicate relationship linked together by what they eat. The largest of the ocean's inhabitants, the Blue Whale, eats very small plankton, which themselves eat even smaller phytoplankton. All of the linkages between predators, grazers, and primary producers in the ocean make up an enormously complicated food web.The base of this food web depends upon phytoplankton, very small photosynthetic organisms which can make their own energy by using energy from the sun. These phytoplankton provide the primary source of the essential nutrients that cycle through our ocean's many food webs. This is called primary productivity, and it is a very good way of measuring the health and abundance of our fisheries.There are many different kinds of phytoplankton in our oceans. [...] One way to study plankton is to collect the plankton using a plankton net to collect samples of macroscopic and microscopic plankton organisms. The net is cast out into the water or trolled behind a boat for a given distance then retrieved. Upon retrieving the net, the contents of the collecting bottle can be removed and the captured plankton can be observed with a microscope. The plankton net will collect both phytoplankton (photosynthetic plankton) and zooplankton (non-photosynthetic plankton and larvae) for observation.In this experiment you will make your own plankton net and use it to collect samples of plankton from different marine or aquatic locations in your local area. You can observe both the abundance (total number of organisms) and diversity (number of different kinds of organisms) of planktonic forms to make conclusions about the productivity and health of each location. In this experiment you will make a plankton net to collect samples of plankton from different locations as an indicator of primary productivity. You can also count the number of phytoplankton (which appear green or brown) compared to zooplankton (which are mostly marine larval forms) and compare. Do the numbers balance, or is there more of one type than the other? What effect do you think this has on productivity cycles?  Food chains are very complex. Find out what types of predators and grazers you have in your area. You can find this information from a field guide or from your local Department of Fish and Game. Can you use this information to construct a food web for your local area?  Some blooms of phytoplankton can be harmful and create an anoxic environment that can suffocate the ecosystem and leave a "Dead Zone" behind. Did you find an excess of brown algae or diatoms? These can be indicators of a harmful algal bloom. Re-visit this location over several weeks to report on an increase or decrease of these types of phytoplankton. Do you think that a harmful algal bloom could be forming in your area? For an experiment that studies the relationship between water quality and algal bloom events, see the Science Buddies project Harmful Algal Blooms in the Chesapeake Bay. 
\end{quote}
\end{spacing}
\end{tiny}
\caption{C-P Pair (Instance $261$ of the original data set.)} 
\label{fig:example}
\end{figure}

\subsection{\cite{gong2018}'s Approach}
The approach by \cite{gong2018} is based on the idea that the longer document in the pair is reduced to a set of \emph{topics} which capture the essence of the document in a way that eliminates the effect of a potential length difference. 
In order to overcome the vocabulary mismatch, these topics are not based on \emph{words} and their distributions (as in LSI (\cite{deerwester1990}) or LDA (\cite{blei2003})), but on word embedding vectors.
Then, basically, matching is done by measuring the cosine similarity between the topic vectors and the short document words.
\cite{gong2018} motivate their approach mainly with the length mismatch argument, which they claim makes approaches relying on document representations (incl.\ vector averaging) unsuitable. 
Accordingly, they use Doc2Vec (\cite{le2014}) as one of their baselines, and show that its performance is inferior to their method. 
They do not, however, provide a much simpler averaging-based baseline. 
As a second baseline, they use Word Mover's Distance (\cite{kusner2015}), which is based on word-level distances, rather than distance of global document representations, but which also fails to be competitive with their topic-based method.
\cite{gong2018} use two different sets of word embeddings: One (topic\_wiki) was trained on a full English Wikipedia dump, the other (wiki\_science) on a smaller subset of the former dump which only contained science articles. 

\section{Our Method}
\label{sec:ourapproach}
We develop our method as a simple alternative to that of \cite{gong2018}. 
We aim at comparable or better classification performance, but with a simpler model. 
Also, we design the method in such a way that it provides human-interpretable results in an efficient way.
\noindent
One common way to compute the similarity of two documents (i.e.\ word sequences) $c$ and $p$ is to average over the word embeddings for each sequence first, and to compute the cosine similarity between the two averages afterwards. 
In the first step, weighting can be applied by multiplying a vector with the TF, IDF, or TF*IDF score of its pertaining word.
We implement this standard measure (\textbf{AVG\_COS\_SIM}) as a baseline for both our method and for the method by \cite{gong2018}. It yields a single scalar similarity score.
\noindent
The core idea of our alternative method is to turn the above process upside down, by computing the cosine similarity of \emph{selected} pairs of words from $c$ and $p$ first, and to average over the similarity scores afterwards (cf.\ also Section \ref{sec:related}).
More precisely, we implement a measure \textbf{TOP\_n\_COS\_SIM\_AVG} as the average of the $n$ highest pairwise cosine similarities of the $n$ top-ranking words in $c$ and $p$.
Ranking, again, is done by TF, IDF, and TF*IDF. For each ranking, we take the top-ranking $n$ words from $c$ and $p$, compute $n \times n$ similarities, rank by decreasing similarity, and average over the top $n$ similarities. 
This measure yields both a scalar similarity score and a list of $<c_x, p_y, sim>$ tuples, which represent the qualitative aspects of $c$ and $p$ on which the similarity score is based.

\section{Experiments}
\label{sec:experiments}
\paragraph{Setup}
All experiments are based on off-the-shelf word-level resources: 
We employ WOMBAT (\cite{mueller2018}) for easy access to the 840B GloVe (\cite{pennington2014}) and the GoogleNews\footnote{\url{https://code.google.com/archive/p/word2vec/}} Word2Vec (\cite{mikolov13c}) embeddings. 
These embedding resources, while slightly outdated, are still widely used.
However, they cannot handle out-of-vocabulary tokens due to their fixed, word-level lexicon.
Therefore, we also use a pretrained English fastText model\footnote{\url{https://dl.fbaipublicfiles.com/fasttext/vectors-crawl/cc.en.300.bin.gz}} (\cite{bojanowski2017,grave2018}), which also includes subword information.
IDF weights for approx.\ $12$ mio.\ different words were obtained from the English Wikipedia dump provided by the Polyglot project (\cite{alrfou2013}).
All resources are case-sensitive, i.e.\ they might contain different entries for words that only differ in case (cf.\ Section \ref{sec:detailanalysis}).

\noindent
We run experiments in different setups, varying both the input representation (GloVe vs.\ Google vs.\ fastText embeddings, $\pm$ TF-weighting, and $\pm$ IDF-weighting) for concepts and projects, and the extent to which concept descriptions are used:
For the latter, \textbf{Label} means only the concept \emph{label} (first and second row in the example), \textbf{Description} means only the textual \emph{description} of the concept, and \textbf{Both} means the concatenation of \textbf{Label} and \textbf{Description}.
For the projects, we always use both label and description. For the project descriptions, we extract only the last column of the original file (CONTENT), and remove user comments and some boiler-plate. 
Each instance in the resulting data set is a tuple of $<c, p, label>$, where $c$ and $p$ are bags of words, with case preserved and function words\footnote{We use the list provided by \cite{gong2018}, with an additional entry for \emph{cannot}.} removed, and $label$ is either $0$ or $1$. 

\paragraph{Parameter Tuning}
Our method is unsupervised, but we need to define a threshold parameter which controls the \emph{minimum} similarity that a concept and a project description should have in order to be considered a match.
Also, the TOP\_n\_COS\_SIM\_AVG measure has a parameter $n$ which controls how many ranked words are used from $c$ and $p$, and how many similarity scores are averaged to create the final score. 
Parameter tuning experiments were performed on a random subset of $20\%$ of our data set ($54$\% positive). Note that \cite{gong2018} used only $10\%$ of their $537$ instances data set as tuning data.
The tuning data results of the best-performing parameter values for each setup can be found in Tables \ref{tab:tuningresults_avg} and \ref{tab:tuningresults_sim_avg}. 
The top F scores per type of concept input (Label, Description, Both) are given in \textbf{bold}.
For AVG\_COS\_SIM and TOP\_n\_COS\_SIM\_AVG, we determined the threshold values (T) on the tuning data by doing a simple $.005$ step search over the range from $0.3$ to $1.0$.
For TOP\_n\_COS\_SIM\_AVG, we additionally varied the value of $n$ in steps of $2$ from $2$ to $30$.

\paragraph{Results}
The top \textbf{tuning data} scores for AVG\_COS\_SIM (Table \ref{tab:tuningresults_avg}) show that the Google embeddings with TF*IDF weighting yield the top F score for all three concept input types ($.881$ - $.945$). 
Somewhat expectedly, the best overall F score ($.945$) is produced in the setting \textbf{Both}, which provides the most information.
Actually, this is true for all four weighting schemes for both GloVe and Google, while fastText consistently yields its top F scores ($.840$ - $.911$) in the \textbf{Label} setting, which provides the least information.
Generally, the level of performance of the simple baseline measure AVG\_COS\_SIM on this data set is rather striking. 

\begin{table}[h]
\centering
\begin{scriptsize}
\renewcommand{\tabcolsep}{2pt}
\begin{tabular}{ccc|l|l|l|l|l|l|l|l|l|l|l|l}
\multicolumn{3}{r|}{\textbf{Concept Input $\rightarrow$}} & \multicolumn{4}{c|}{\textbf{Label}} & \multicolumn{4}{c|}{\textbf{Description}} & \multicolumn{4}{c}{\textbf{Both}} \\ 
\textbf{Embeddings}              & \textbf{TF} & \textbf{IDF} & \multicolumn{1}{c}{\textbf{T}} & \multicolumn{1}{c}{\textbf{P}} & \multicolumn{1}{c}{\textbf{R}} & \multicolumn{1}{c|}{\textbf{F}}  & \multicolumn{1}{c}{\textbf{T}} & \multicolumn{1}{c}{\textbf{P}} & \multicolumn{1}{c}{\textbf{R}} & \multicolumn{1}{c|}{\textbf{F}} & \multicolumn{1}{c}{\textbf{T}} & \multicolumn{1}{c}{\textbf{P}} & \multicolumn{1}{c}{\textbf{R}} & \multicolumn{1}{c}{\textbf{F}}\\
\hline
\multirow{4}{*}{\textbf{GloVe}}      & \textbf{-}  & \textbf{-}   & $.635$  & $.750$  & $.818$  & $.783$          & $.720$  & $.754$ & $.891$ & $.817$          & $.735$  & $.765$           & $.945$           & $.846$\\
                                     & \textbf{+}  & \textbf{-}   & $.640$  & $.891$  & $.745$  & $.812$          & $.700$  & $.831$ & $.891$ & $.860$          & $.690$  & $.813$           & $.945$           & $.874$\\
                                     & \textbf{-}  & \textbf{+}   & $.600$  & $.738$  & $.873$  & $.800$          & $.670$  & $.746$ & $.909$ & $.820$          & $.755$  & $.865$           & $.818$           & $.841$\\
                                     & \textbf{+}  & \textbf{+}   & $.605$  & $.904$  & $.855$  & $.879$          & $.665$  & $.857$ & $.873$ & $.865$          & $.715$  & $.923$           & $.873$           & $.897$\\
\cline{1-15}
\multirow{4}{*}{\textbf{Google}}     & \textbf{-}  & \textbf{-}   & $.440$  & $.813$  & $.945$  & $.874$          & $.515$  & $.701$ & $.982$ & $.818$          & $.635$  & $.920$           & $.836$           & $.876$\\
                                     & \textbf{+}  & \textbf{-}   & $.445$  & $.943$  & $.909$  & $\textbf{.926}$ & $.540$  & $.873$ & $.873$ & $.873$          & $.565$  & $.927$           & $.927$           & $.927$\\
                                     & \textbf{-}  & \textbf{+}   & $.435$  & $.839$  & $.945$  & $.889$          & $.520$  & $.732$ & $.945$ & $.825$          & $.590$  & $.877$           & $.909$           & $.893$\\
                                     & \textbf{+}  & \textbf{+}   & $.430$  & $.943$  & $.909$  & $\textbf{.926}$ & $.530$  & $.889$ & $.873$ & $\textbf{.881}$ & $.545$  & $.945$           & $.945$           & $\textbf{.945}$\\
\cline{1-15}
\multirow{4}{*}{\textbf{fastText}}   & \textbf{-}  & \textbf{-}   & $.440$  & $.781$  & $.909$  & $.840$          & $.555$  & $.708$ & $.927$ & $.803$          & $.615$  & $.778$           & $.891$           & $.831$\\
                                     & \textbf{+}  & \textbf{-}   & $.435$  & $.850$  & $.927$  & $.887$          & $.520$  & $.781$ & $.909$ & $.840$          & $.530$  & $.803$           & $.964$           & $.876$\\
                                     & \textbf{-}  & \textbf{+}   & $.435$  & $.850$  & $.927$  & $.887$          & $.525$  & $.722$ & $.945$ & $.819$          & $.600$  & $.820$           & $.909$           & $.862$\\
                                     & \textbf{+}  & \textbf{+}   & $.420$  & $.895$  & $.927$  & $.911$          & $.505$  & $.803$ & $.891$ & $.845$          & $.520$  & $.833$           & $.909$           & $.870$\\

\cline{1-15}
\end{tabular} 
\caption{Tuning Data Results \textbf{AVG\_COS\_SIM}. Top F per Concept Input Type in \textbf{Bold}.}
\label{tab:tuningresults_avg}
\end{scriptsize}
\end{table}

\noindent
For TOP\_n\_COS\_SIM\_AVG, the \textbf{tuning data} results (Table \ref{tab:tuningresults_sim_avg}) are somewhat more varied:
First, there is no single best performing set of embeddings: Google yields the best F score for the \textbf{Label} setting ($.953$), while GloVe (though only barely) leads in the \textbf{Description} setting ($.912$).
This time, it is fastText which produces the best F score in the \textbf{Both} setting, which is also the best overall \textbf{tuning data} F score for TOP\_n\_COS\_SIM\_AVG ($.954$). 
While the difference to the Google result for \textbf{Label} is only minimal, it is striking that the best overall score is again produced using the 'richest' setting, i.e.\ the one involving both TF and IDF weighting and the most informative input. 

\begin{table}[h]
\centering
\begin{scriptsize}
\renewcommand{\tabcolsep}{2pt}
\begin{tabular}{ccc|l|l|l|l|l|l|l|l|l|l|l|l}
\multicolumn{3}{r|}{\textbf{Concept Input $\rightarrow$}} & \multicolumn{4}{c|}{\textbf{Label}} & \multicolumn{4}{c|}{\textbf{Description}} & \multicolumn{4}{c}{\textbf{Both}} \\ 
\textbf{Embeddings}              & \textbf{TF} & \textbf{IDF} & \multicolumn{1}{c}{\textbf{T/n}} & \multicolumn{1}{c}{\textbf{P}} & \multicolumn{1}{c}{\textbf{R}} & \multicolumn{1}{c|}{\textbf{F}}  & \multicolumn{1}{c}{\textbf{T/n}} & \multicolumn{1}{c}{\textbf{P}} & \multicolumn{1}{c}{\textbf{R}} & \multicolumn{1}{c|}{\textbf{F}} & \multicolumn{1}{c}{\textbf{T/n}} & \multicolumn{1}{c}{\textbf{P}} & \multicolumn{1}{c}{\textbf{R}} & \multicolumn{1}{c}{\textbf{F}}\\
\hline
\multirow{3}{*}{\textbf{GloVe}}      & \textbf{+} & \textbf{-} & $.365/ 6$ & $.797$ & $.927$ & $.857$          & $.690/14$ & $.915$ & $.782$ & $.843$          & $.675/16$ & $.836$ & $.927$ & $.879$\\
                                     & \textbf{-} & \textbf{+} & $.300/30$ & $.929$ & $.236$ & $.377$          & $.300/30$ & $.806$ & $.455$ & $.581$          & $.300/30$ & $.778$ & $.636$ & $.700$\\
                                     & \textbf{+} & \textbf{+} & $.330/ 6$ & $.879$ & $.927$ & $.903$          & $.345/ 6$ & $.881$ & $.945$ & $\textbf{.912}$ & $.345/ 6$ & $.895$ & $.927$ & $.911$\\
\cline{1-15}
\multirow{3}{*}{\textbf{Google}}     & \textbf{+} & \textbf{-} & $.345/22$ & $.981$ & $.927$ & $\textbf{.953}$ & $.480/16$ & $.895$ & $.927$ & $.911$          & $.520/16$ & $.912$ & $.945$ & $.929$\\
                                     & \textbf{-} & \textbf{+} & $.300/30$ & $1.00$ & $.345$ & $.514$          & $.300/ 8$ & $1.00$ & $.345$ & $.514$          & $.300/30$ & $1.00$ & $.600$ & $.750$\\
                                     & \textbf{+} & \textbf{+} & $.300/10$ & $1.00$ & $.509$ & $.675$          & $.300/14$ & $.972$ & $.636$ & $.769$          & $.350/22$ & $1.00$ & $.836$ & $.911$\\
\cline{1-15}
\multirow{3}{*}{\textbf{fastText}}   & \textbf{+} & \textbf{-} & $.415/22$ & $.980$ & $.873$ & $.923$          & $.525/14$ & $.887$ & $.855$ & $.870$          & $.535/20$ & $.869$ & $.964$ & $.914$\\
                                     & \textbf{-} & \textbf{+} & $.350/24$ & $1.00$ & $.309$ & $.472$          & $.300/30$ & $1.00$ & $.382$ & $.553$          & $.300/28$ & $1.00$ & $.673$ & $.804$\\
                                     & \textbf{+} & \textbf{+} & $.300/20$ & $1.00$ & $.800$ & $.889$          & $.300/10$ & $.953$ & $.745$ & $.837$          & $.310/14$ & $.963$ & $.945$ & $\textbf{.954}$\\

\cline{1-15}
\end{tabular} 
\caption{Tuning Data Results \textbf{TOP\_n\_COS\_SIM\_AVG}. Top F per Concept Input Type in \textbf{Bold}.}
\label{tab:tuningresults_sim_avg}
\end{scriptsize}
\end{table}

\noindent
We then selected the best performing parameter settings for every concept input and ran experiments on the held-out \textbf{test data}.
Since the original data split used by \cite{gong2018} is unknown, we cannot exactly replicate their settings, 
but we also perform ten runs using randomly selected $10\%$ of our $408$ instances test data set, and report average P, R, F, and standard deviation. 
The results can be found in Table \ref{tab:results}. 
For comparison, the two top rows provide the best results of \cite{gong2018}. 

\begin{table}[h]
\centering
\begin{scriptsize}
\renewcommand{\tabcolsep}{5pt}
\begin{tabular}{llllll|cc|cc|cc}
\multicolumn{6}{c|}{\textbf{}} & \multicolumn{2}{c}{\textbf{P}} & \multicolumn{2}{c}{\textbf{R}} & \multicolumn{2}{c}{\textbf{F}} \\
\hline
\multirow{2}{*}{\textbf{\cite{gong2018}}}    & \multicolumn{5}{l|}{topic\_science}  & $.758$ & $\pm .012$ & $.885$ & $\pm .071$ & $.818$ & $\pm .028$\\
                                             & \multicolumn{5}{l|}{topic\_wiki}     & $.750$ & $\pm .009$ & $.842$ & $\pm .010$ & $.791$ & $\pm .007$\\
\hline
\hline
\textbf{Method}                                 & \textbf{Embeddings} & \multicolumn{2}{c}{\textbf{Settings}} & \multicolumn{1}{c}{\textbf{T/n}}       & \textbf{Conc.\ Input} & &&&&&\\
\multirow{3}{*}{\textbf{AVG\_COS\_SIM}}         & Google              & +TF & +IDF                            & $.515$    & Label       & $.939$ & $\pm .043$ & $.839$ & $\pm .067$ & $.884$          & $\pm .038$\\
                                                & Google              & +TF & +IDF                            & $.520$    & Description & $.870$ & $\pm .068$ & $.834$ & $\pm .048$ & $.849$          & $\pm .038$\\
                                                & Google              & +TF & +IDF                            & $.545$    & Both        & $.915$ & $\pm .040$ & $.938$ & $\pm .047$ & $\textbf{.926}$ & $\pm .038$\\
\hline
\multirow{3}{*}{\textbf{TOP\_n\_COS\_SIM\_AVG}} & Google              & +TF & -IDF                            & $.345/22$ & Label       & $.854$ & $\pm .077$ & $.861$ & $\pm .044$ & $.856$          & $\pm .054$\\
                                                & GloVe               & +TF & +IDF                            & $.345/6 $ & Description & $.799$ & $\pm .063$ & $.766$ & $\pm .094$ & $.780$          & $\pm .068$\\
                                                & fastText            & +TF & +IDF                            & $.310/14$ & Both        & $.850$ & $\pm .059$ & $.918$ & $\pm .049$ & $\textbf{.881}$ & $\pm .037$\\
\hline
\end{tabular} 
\caption{Test Data Results}
\label{tab:results}
\end{scriptsize}
\end{table}

\noindent
The first interesting finding is that the AVG\_COS\_SIM measure again performs very well: 
In all three settings, it beats both the system based on general-purpose embeddings (topic\_wiki) and the one that is adapted to the science domain (topic\_science), with again the \textbf{Both} setting yielding the best overall result ($.926$).
Note that our \textbf{Both} setting is probably the one most similar to the concept input used by \cite{gong2018}. 
This result corroborates our findings on the tuning data, and clearly contradicts the (implicit) claim made by \cite{gong2018} regarding the infeasibility of document-level matching for documents of different lengths.
\noindent
The second, more important finding is that our proposed TOP\_n\_COS\_SIM\_AVG measure is also very competitive, as it also outperforms both systems by \cite{gong2018} in two out of three settings. It only fails in the setting using only the \textbf{Description} input.\footnote{Remember that this setup was only minimally superior ($.001$ F score) to the next best one on the tuning data.} 
This is the more important as we exclusively employ off-the-shelf, general-purpose embeddings, while \cite{gong2018} reach their best results with a much more sophisticated system and with embeddings that were custom-trained for the science domain.
Thus, while the performance of our proposed TOP\_n\_COS\_SIM\_AVG method is superior to the approach by \cite{gong2018}, it is itself outperformed by the 'baseline' AVG\_COS\_SIM method with appropriate weighting.
However, apart from raw classification performance, our method also aims at providing human-interpretable information on how a classification was done. 
In the next section, we perform a detail analysis on a selected setup.

\section{Detail Analysis}
\label{sec:detailanalysis}
The similarity-labelled word pairs from concept and project description which are selected during classification with the TOP\_n\_COS\_ SIM\_AVG measure provide a way to qualitatively evaluate the basis on which each similarity score was computed. We see this as an advantage over average-based comparison (like AVG\_COS\_SIM), since it provides a means to check the plausibility of the decision. 
Here, we are mainly interested in the overall best result, so we perform a detail analysis on the best-performing \textbf{Both} setting only (fastText, TF*IDF weighting, $T=.310$, $n=14$).
Since the \emph{Concept-Project matching} task is a binary classification task, its performance can be qualitatively analysed by providing examples for instances that were classified correctly (True Positive (TP) and True Negative (TN)) or incorrectly (False Positive (FP) and False Negative (FN)).

\noindent
Table \ref{tab:detailone} shows the concept and project words from selected instances (one TP, FP, TN, and FN case each) of the tuning data set.
Concept and project words are ordered alphabetically, with concept words appearing more than once being grouped together.
According to the selected setting, the number of word pairs is $n=14$. 
The bottom line in each column provides the average similarity score as computed by the TOP\_n\_COS\_SIM\_AVG measure. 
This value is compared against the threshold $T=.310$.
The similarity is higher than $T$ in the TP and FP cases, and lower otherwise. 
Without going into too much detail, it can be seen that the selected words provide a reasonable idea of the gist of the two documents.
Another observation relates to the effect of using unstemmed, case-sensitive documents as input:
the top-ranking words often contain inflectional variants (e.g.\ \emph{enzyme} and \emph{enzymes}, \emph{level} and \emph{levels} in the example), and words differing in case only can also be found. 
Currently, these are treated as distinct (though semantically similar) words, mainly out of compatibility with the pretrained GloVe and Google embeddings. 
However, since our method puts a lot of emphasis on individual words, in particular those coming from the shorter of the two documents (the \emph{concept}), results might be improved by somehow merging these words (and their respective embedding vectors) (see Section \ref{sec:conclusion}).

\begin{table}
\centering
\begin{footnotesize}
\renewcommand{\tabcolsep}{3pt}
\begin{tabular}{lll|lll|lll|lll}
\multicolumn{3}{c}{TP ($\textbf{.447} > .310$)} & \multicolumn{3}{c}{FP ($\textbf{.367} > .310$)}  & \multicolumn{3}{c}{TN ($\textbf{.195} < .310$)} & \multicolumn{3}{c}{FN ($\textbf{.278} < .310$)} \\
\hline
\multicolumn{1}{c}{\textbf{Concept}} & \multicolumn{1}{c}{\textbf{Project}} & \multirow{2}{*}{\textbf{Sim}} & \multicolumn{1}{c}{\textbf{Concept}} & \multicolumn{1}{c}{\textbf{Project}} & \multirow{2}{*}{\textbf{Sim}} & \multicolumn{1}{c}{\textbf{Concept}} & \multicolumn{1}{c}{\textbf{Project}} & \multirow{2}{*}{\textbf{Sim}} & \multicolumn{1}{c}{\textbf{Concept}} & \multicolumn{1}{c}{\textbf{Project}} & \multirow{2}{*}{\textbf{Sim}}\\
\multicolumn{1}{c}{\textbf{Word}}    & \multicolumn{1}{c}{\textbf{Word}}    &                               & \multicolumn{1}{c}{\textbf{Word}}    & \multicolumn{1}{c}{\textbf{Word}}    &                                & \multicolumn{1}{c}{\textbf{Word}}    & \multicolumn{1}{c}{\textbf{Word}}    &                               & \multicolumn{1}{c}{\textbf{Word}}    & \multicolumn{1}{c}{\textbf{Word}}    & \\
\hline
cells     & enzymes                  & $.438$   & co-evolution   & dynamic   & $.299$ & energy   & allergy         & $.147$ & area    & water         & $.277$\\
cells     & genes                    & $.427$   & continual      & dynamic   & $.296$ & energy   & juice           & $.296$ & climate & water         & $.269$\\
molecules & DNA                      & $.394$   & delicate       & detail    & $.306$ & energy   & leavening       & $.186$ & earth   & copper        & $.254$\\ 
molecules & enzyme                   & $.445$   & delicate       & dynamic   & $.326$ & energy   & substitutes     & $.177$ & earth   & metal         & $.277$\\ 
molecules & enzymes                  & $.533$   & delicate       & texture   & $.379$ & surface  & average         & $.212$ & earth   & metals        & $.349$\\ 
molecules & gene                     & $.369$   & dynamic        & dynamic   & $1.00$ & surface  & baking          & $.216$ & earth   & water         & $.326$\\ 
molecules & genes                    & $.471$   & dynamic        & image     & $.259$ & surface  & egg             & $.178$ & extent  & concentration & $.266$\\ 
multiple  & different                & $.550$   & dynamic        & range     & $.377$ & surface  & leavening       & $.158$ & range   & concentration & $.255$\\ 
organisms & enzyme                   & $.385$   & dynamic        & texture   & $.310$ & surface  & thickening      & $.246$ & range   & ppm           & $.237$\\ 
organisms & enzymes                  & $.512$   & surface        & level     & $.323$ & transfer & baking          & $.174$ & systems & metals        & $.243$\\ 
organisms & genes                    & $.495$   & surface        & texture   & $.383$ & transfer & substitute      & $.192$ & systems & solution      & $.275$\\ 
organs    & enzymes                  & $.372$   & surface        & tiles     & $.321$ & transfer & substitutes     & $.157$ & typical & heavy         & $.299$\\ 
tissues   & enzymes                  & $.448$   & systems        & dynamic   & $.272$ & warms    & baking          & $.176$ & weather & heavy         & $.248$\\ 
tissues   & genes                    & $.414$   & systems        & levels    & $.286$ & warms    & thickening      & $.214$ & weather & water         & $.308$\\
\hline
          & \multicolumn{1}{r}{Avg.\ Sim} & $\textbf{.447}$   & & \multicolumn{1}{r}{Avg.\ Sim} & $\textbf{.367}$ & & \multicolumn{1}{r}{Avg.\ Sim} & $\textbf{.195}$ & & \multicolumn{1}{r}{Avg.\ Sim} & $\textbf{.278}$ \\
\end{tabular}
\end{footnotesize}
\label{tab:detailone}
\caption{\textbf{TOP\_n\_COS\_SIM\_AVG} Detail Results of Best-performing fastText Model on \textbf{Both}.}
\end{table}

\section{Related Work}
\label{sec:related}
While in this paper we apply our method to the \emph{Concept-Project matching} task only, the underlying task of matching text sequences to each other is much more general.
Many existing approaches follow the so-called \emph{compare-aggregate} framework (\cite{wang2017}). As the name suggests, these approaches collect the results of element-wise matchings (\emph{comparisons}) first, and
create the final result by aggregating these results later. Our method can be seen as a variant of \emph{compare-aggregate} which is characterized by extremely simple methods for comparison (cosine vector similarity) and aggregation (averaging). Other approaches, like \cite{he2016} and \cite{wang2017}, employ much more elaborated supervised neural networks methods. 
Also, on a simpler level, the idea of averaging similarity scores (rather than scoring averaged representations) is not new:
\cite{camacho-collados2016} use the average of pairwise word similarities to compute their \emph{compactness score}.

\section{Conclusion and Future Work}
\label{sec:conclusion}
We presented a simple method for semantic matching of documents from heterogeneous collections as a solution to the \emph{Concept-Project matching} task by \cite{gong2018}. 
Although much simpler, our method clearly outperformed the original system in most input settings. 
Another result is that, contrary to the claim made by \cite{gong2018}, the standard averaging approach does indeed work very well even for heterogeneous document collections, if appropriate weighting is applied. 
Due to its simplicity, we believe that our method can also be applied to other text matching tasks, including more 'standard' ones which do not necessarily involve \textbf{heterogeneous} document collections. 
This seems desirable because our method offers additional transparency by providing not only a similarity score, but also the subset of words on which the similarity score is based.
Future work includes detailed error analysis, and exploration of methods to combine complementary information about (grammatically or orthographically) related words from word embedding resources.
Also, we are currently experimenting with a pretrained ELMo (\cite{peters2018}) model as another word embedding resource. 
ELMo takes word embeddings a step further by dynamically creating \emph{contextualized} vectors from input word sequences (normally sentences). 
Our initial experiments have been promising, but since 
ELMo tends to yield \emph{different}, context-dependent vectors for the \emph{same} word in the \emph{same} document, ways have still to be found to combine them into single, document-wide vectors, without (fully) sacrificing their context-awareness.

\noindent
The code used in this paper is available at \url{https://github.com/nlpAThits/TopNCosSimAvg}. 

\vspace{10pt}
\noindent
\textbf{Acknowledgements}
\noindent
The research described in this paper was funded by the Klaus Tschira Foundation. We thank the anonymous reviewers for their useful comments and suggestions.

\bibliography{acl2019}

\begin{thebibliography}{}

\bibitem[\protect\citeauthoryear{Al-Rfou, Perozzi, and Skiena}{Al-Rfou
  et~al.}{2013}]{alrfou2013}
Al-Rfou, R., B.~Perozzi, and S.~Skiena (2013, August).
\newblock Polyglot: Distributed word representations for multilingual nlp.
\newblock In {\em Proceedings of the Seventeenth Conference on Computational
  Natural Language Learning}, Sofia, Bulgaria, pp.\  183--192. Association for
  Computational Linguistics.

\bibitem[\protect\citeauthoryear{Blei, Ng, and Jordan}{Blei
  et~al.}{2003}]{blei2003}
Blei, D.~M., A.~Y. Ng, and M.~I. Jordan (2003).
\newblock Latent dirichlet allocation.
\newblock {\em Journal of Machine Learning Research\/}~{\em 3}, 993--1022.

\bibitem[\protect\citeauthoryear{Blooma and Kurian}{Blooma and
  Kurian}{2011}]{blooma2011}
Blooma, M.~J. and J.~C. Kurian (2011).
\newblock Research issues in community based question answering.
\newblock In P.~B. Seddon and S.~Gregor (Eds.), {\em Pacific Asia Conference on
  Information Systems, {PACIS} 2011: Quality Research in Pacific Asia,
  Brisbane, Queensland, Australia, 7-11 July 2011}, pp.\ ~29. Queensland
  University of Technology.

\bibitem[\protect\citeauthoryear{Bojanowski, Grave, Joulin, and
  Mikolov}{Bojanowski et~al.}{2017}]{bojanowski2017}
Bojanowski, P., E.~Grave, A.~Joulin, and T.~Mikolov (2017).
\newblock Enriching word vectors with subword information.
\newblock {\em {TACL}\/}~{\em 5}, 135--146.

\bibitem[\protect\citeauthoryear{Camacho{-}Collados and
  Navigli}{Camacho{-}Collados and Navigli}{2016}]{camacho-collados2016}
Camacho{-}Collados, J. and R.~Navigli (2016).
\newblock Find the word that does not belong: {A} framework for an intrinsic
  evaluation of word vector representations.
\newblock In {\em Proceedings of the 1st Workshop on Evaluating Vector-Space
  Representations for NLP, RepEval@ACL 2016, Berlin, Germany, August 2016},
  pp.\  43--50. Association for Computational Linguistics.

\bibitem[\protect\citeauthoryear{Deerwester, Dumais, Landauer, Furnas, and
  Harshman}{Deerwester et~al.}{1990}]{deerwester1990}
Deerwester, S.~C., S.~T. Dumais, T.~K. Landauer, G.~W. Furnas, and R.~A.
  Harshman (1990).
\newblock Indexing by latent semantic analysis.
\newblock {\em {JASIS}\/}~{\em 41\/}(6), 391--407.

\bibitem[\protect\citeauthoryear{Feng, Xiang, Glass, Wang, and Zhou}{Feng
  et~al.}{2015}]{feng2015}
Feng, M., B.~Xiang, M.~R. Glass, L.~Wang, and B.~Zhou (2015).
\newblock Applying deep learning to answer selection: {A} study and an open
  task.
\newblock In {\em 2015 {IEEE} Workshop on Automatic Speech Recognition and
  Understanding, {ASRU} 2015, Scottsdale, AZ, USA, December 13-17, 2015}, pp.\
  813--820. {IEEE}.

\bibitem[\protect\citeauthoryear{Gong, Sakakini, Bhat, and Xiong}{Gong
  et~al.}{2018}]{gong2018}
Gong, H., T.~Sakakini, S.~Bhat, and J.~Xiong (2018).
\newblock Document similarity for texts of varying lengths via hidden topics.
\newblock In {\em Proceedings of the 56th Annual Meeting of the Association for
  Computational Linguistics (Volume 1: Long Papers)}, pp.\  2341--2351.
  Association for Computational Linguistics.

\bibitem[\protect\citeauthoryear{Grave, Bojanowski, Gupta, Joulin, and
  Mikolov}{Grave et~al.}{2018}]{grave2018}
Grave, E., P.~Bojanowski, P.~Gupta, A.~Joulin, and T.~Mikolov (2018).
\newblock Learning word vectors for 157 languages.
\newblock In N.~Calzolari, K.~Choukri, C.~Cieri, T.~Declerck, S.~Goggi,
  K.~Hasida, H.~Isahara, B.~Maegaard, J.~Mariani, H.~Mazo, A.~Moreno, J.~Odijk,
  S.~Piperidis, and T.~Tokunaga (Eds.), {\em Proceedings of the Eleventh
  International Conference on Language Resources and Evaluation, {LREC} 2018,
  Miyazaki, Japan, May 7-12, 2018.} European Language Resources Association
  {(ELRA)}.

\bibitem[\protect\citeauthoryear{He and Lin}{He and Lin}{2016}]{he2016}
He, H. and J.~J. Lin (2016).
\newblock Pairwise word interaction modeling with deep neural networks for
  semantic similarity measurement.
\newblock In K.~Knight, A.~Nenkova, and O.~Rambow (Eds.), {\em {NAACL} {HLT}
  2016, The 2016 Conference of the North American Chapter of the Association
  for Computational Linguistics: Human Language Technologies, San Diego
  California, USA, June 12-17, 2016}, pp.\  937--948. The Association for
  Computational Linguistics.

\bibitem[\protect\citeauthoryear{Kusner, Sun, Kolkin, and Weinberger}{Kusner
  et~al.}{2015}]{kusner2015}
Kusner, M.~J., Y.~Sun, N.~I. Kolkin, and K.~Q. Weinberger (2015).
\newblock From word embeddings to document distances.
\newblock In F.~R. Bach and D.~M. Blei (Eds.), {\em Proceedings of the 32nd
  International Conference on Machine Learning, {ICML} 2015, Lille, France,
  6-11 July 2015}, Volume~37 of {\em {JMLR} Workshop and Conference
  Proceedings}, pp.\  957--966. JMLR.org.

\bibitem[\protect\citeauthoryear{Le and Mikolov}{Le and Mikolov}{2014}]{le2014}
Le, Q.~V. and T.~Mikolov (2014).
\newblock Distributed representations of sentences and documents.
\newblock In {\em Proceedings of the 31th International Conference on Machine
  Learning, {ICML} 2014, Beijing, China, 21-26 June 2014}, Volume~32 of {\em
  {JMLR} Workshop and Conference Proceedings}, pp.\  1188--1196. JMLR.org.

\bibitem[\protect\citeauthoryear{Mikolov, Sutskever, Chen, Corrado, and
  Dean}{Mikolov et~al.}{2013}]{mikolov13c}
Mikolov, T., I.~Sutskever, K.~Chen, G.~S. Corrado, and J.~Dean (2013).
\newblock Distributed representations of words and phrases and their
  compositionality.
\newblock In {\em Proceedings of Advances in Neural Information Processing
  Systems 26. {\em Lake Tahoe, Nev., 5--8 December 2013}}, pp.\  3111--3119.

\bibitem[\protect\citeauthoryear{M{\"{u}}ller and Strube}{M{\"{u}}ller and
  Strube}{2018}]{mueller2018}
M{\"{u}}ller, M. and M.~Strube (2018).
\newblock Transparent, efficient, and robust word embedding access with
  {WOMBAT}.
\newblock In D.~Zhao (Ed.), {\em {COLING} 2018, The 27th International
  Conference on Computational Linguistics: System Demonstrations, Santa Fe, New
  Mexico, August 20-26, 2018}, pp.\  53--57. Association for Computational
  Linguistics.

\bibitem[\protect\citeauthoryear{Pennington, Socher, and Manning}{Pennington
  et~al.}{2014}]{pennington2014}
Pennington, J., R.~Socher, and C.~D. Manning (2014).
\newblock Glove: Global vectors for word representation.
\newblock In A.~Moschitti, B.~Pang, and W.~Daelemans (Eds.), {\em Proceedings
  of the 2014 Conference on Empirical Methods in Natural Language Processing,
  {EMNLP} 2014, October 25-29, 2014, Doha, Qatar, {A} meeting of SIGDAT, a
  Special Interest Group of the {ACL}}, pp.\  1532--1543. {ACL}.

\bibitem[\protect\citeauthoryear{Peters, Neumann, Iyyer, Gardner, Clark, Lee,
  and Zettlemoyer}{Peters et~al.}{2018}]{peters2018}
Peters, M.~E., M.~Neumann, M.~Iyyer, M.~Gardner, C.~Clark, K.~Lee, and
  L.~Zettlemoyer (2018).
\newblock Deep contextualized word representations.
\newblock In M.~A. Walker, H.~Ji, and A.~Stent (Eds.), {\em Proceedings of the
  2018 Conference of the North American Chapter of the Association for
  Computational Linguistics: Human Language Technologies, {NAACL-HLT} 2018, New
  Orleans, Louisiana, USA, June 1-6, 2018, Volume 1 (Long Papers)}, pp.\
  2227--2237. Association for Computational Linguistics.

\bibitem[\protect\citeauthoryear{Wang and Jiang}{Wang and
  Jiang}{2017}]{wang2017}
Wang, S. and J.~Jiang (2017).
\newblock A compare-aggregate model for matching text sequences.
\newblock In {\em 5th International Conference on Learning Representations,
  {ICLR} 2017, Toulon, France, April 24-26, 2017, Conference Track
  Proceedings}. OpenReview.net.

\end{thebibliography}
\bibliographystyle{chicago}
\end{document}